\documentclass[twocolumn, switch]{article} 

\usepackage{preprint}
\usepackage{algorithm}
\usepackage{algorithmic}
\usepackage{amsmath, amsthm, amssymb, amsfonts}

\usepackage[numbers,square]{natbib}
\usepackage[utf8]{inputenc}	
\usepackage[T1]{fontenc}	
\usepackage{xcolor}		
\usepackage[colorlinks = true,
            linkcolor = purple,
            urlcolor  = blue,
            citecolor = cyan,
            anchorcolor = black]{hyperref}	
\usepackage{booktabs} 		
\usepackage{nicefrac}		
\usepackage{microtype}		
\usepackage{lineno}		
\usepackage{float}			

\usepackage{lipsum}		

\usepackage{newfloat}
\DeclareFloatingEnvironment[name={Supplementary Figure}]{suppfigure}
\usepackage{sidecap}
\sidecaptionvpos{figure}{c}

\usepackage{titlesec}
\titlespacing*{\section}{0pt}{8pt plus 2pt minus 2pt}{1pt plus 1pt minus 1pt}
\titlespacing*{\subsection}{0pt}{5pt plus 2pt minus 1pt}{1pt plus 1pt minus 1pt}
\titlespacing*{\subsubsection}{0pt}{3pt plus 1pt minus 1pt}{0.5pt plus 0.5pt minus 0.5pt}
\usepackage{amsmath,amssymb,amsfonts}
\usepackage{algorithmic}
\usepackage{graphicx}
\usepackage{textcomp}
\usepackage{xcolor}
\usepackage{amsmath}
\usepackage{cases}
\usepackage{bbm}
\usepackage{subfig}
\usepackage{graphicx}
\usepackage{array}
\usepackage{booktabs}
\usepackage{rotating}
\usepackage{makecell}
\usepackage{colortbl}
\usepackage{multirow}
\usepackage{float}
\usepackage{gensymb}
\usepackage{hyperref}
\usepackage{cleveref}
\usepackage{enumitem}
\usepackage{amssymb}
\usepackage{authblk}
\usepackage{xcolor}
\usepackage{colortbl}   
\usepackage{multirow}   
\newcommand{\impv}[1]{{\color{teal}$\downarrow$#1\%}}
\usepackage[table]{xcolor}
\definecolor{bestred}{rgb}{1,.455,.455}
\definecolor{stagrow}{rgb}{1.00,0.97,0.85}
\definecolor{highcol}{rgb}{0.92,0.96,1.00}
\definecolor{stagborder}{rgb}{0.95,0.65,0.15}

\newcommand{\best}[1]{\cellcolor{bestred}\textbf{#1}}

\title{STAG-VIO: Stabilized Prompt-to-Geometry Interface for Robust Dynamic Visual--Inertial Odometry}

\usepackage{eso-pic}
\usepackage{tikz}
\usepackage{xcolor}
\PassOptionsToPackage{colorlinks=true,linkcolor=gray,urlcolor=gray}{hyperref}

\usepackage{titling}
\usepackage{footmisc}
\newcommand{\Author}[3]{
  \textbf{#1}\textsuperscript{#2}%
}

\author{
  \Author{Rui Zhou}{1}{0000-0000-0000-0000} \and
  \Author{Jingbin Liu}{2,\textdagger}{0000-0000-0000-0000} \and
  \Author{Junbin Xie}{1}{0000-0000-0000-0000} \and
  \Author{Jianyu Zhang}{1}{0000-0000-0000-0000} \and
  \Author{Yingze Hu}{1}{0000-0000-0000-0000} \and
  \Author{Jiele Zhao}{1}{0000-0000-0000-0000}%
}

\date{%
  \textsuperscript{1}Electronic Information School, Wuhan University.\\
  \textsuperscript{2}College of Geodesy and Geomatics, Shandong University of Science and
  Technology\\[0.5em]
  \textbf{Project page:} \href{https://stag-vio.github.io/}{https://stag-vio.github.io/}\\[0.5em]
  \textsuperscript{\textdagger}Corresponding author
}

\begin{document}

\twocolumn[ 
  \begin{@twocolumnfalse} 

\maketitle
\thispagestyle{empty}

\begin{abstract}
Dynamic visual--inertial odometry (VIO) requires reliable suppression
of motion-corrupted measurements, yet prior semantic-assisted
approaches depend on category-limited segmenters and degrade under
partial occlusion. Promptable foundation segmentation models offer
category-agnostic dynamic parsing, but their effectiveness in VIO
depends critically on the temporal stability of input prompts---a
factor largely overlooked in existing pipelines. When prompts derived
from raw detection are jittery or intermittent under occlusion, the
resulting masks flicker across frames, destabilizing geometric
estimation. We propose STAG-VIO, which formulates dynamic robustness
as a perception-to-geometry interface stabilization problem. We
introduce uncertainty-adaptive multi-object tracking that models
prompt generation as state estimation with bounded noise adaptation,
producing temporally coherent box prompts. These stabilized prompts
drive a lightweight foundation segmenter whose masks undergo
geometry-oriented morphological refinement to establish conservative
safety margins. A constraint-budget-aware feature redistribution
strategy preserves well-conditioned static measurements when dynamic
regions dominate the view. Experiments on VIODE and OpenLORIS-Scene
show consistent gains over state-of-the-art baselines. Ablation
confirms that prompt stabilization is the single most impactful
component, reducing trajectory error by up to 83\%.
\end{abstract}

  \end{@twocolumnfalse} 
] 
\begingroup
\renewcommand{\thefootnote}{\fnsymbol{footnote}}
\endgroup



\section{INTRODUCTION}
\label{sec:introduction}

Visual--inertial odometry (VIO), which fuses synchronized
camera streams and IMU measurements for ego-motion estimation,
is a foundational capability for autonomous platforms operating
in GPS-denied environments~\cite{orb-slam3,vins-mono,vins-fusion}.
By exploiting complementary visual and inertial cues, modern VIO
systems achieve accurate, drift-controlled pose estimation under a
wide range of motions.
However, real-world scenes are rarely static: pedestrians,
vehicles, and other moving objects frequently dominate the
camera's field of view, while partial occlusions and appearance
variations further complicate visual perception.
Under such highly dynamic conditions, classical static-scene
assumptions are violated, leading to corrupted feature tracks,
degraded geometric constraints, and---in extreme
cases---complete tracking failure.
A natural remedy is to incorporate semantic segmentation to
suppress dynamic regions before geometric
estimation~\cite{dsslam2018,dynaslam2018,dyna-vins,Dynamic-VINS},
yet this line of work faces a systematic bottleneck that has
received insufficient attention.

Prior semantic-assisted VIO
systems~\cite{dsslam2018,dyna-vins,dynaslam2018,Dynamic-VINS},
rely on segmentation networks trained on fixed label sets
(\emph{e.g.}, SegNet~\cite{SegNet},
Mask~R-CNN~\cite{Mask_rcnn}), which limits their ability to
generalize to unseen dynamic object categories.
Recent promptable foundation models such as
SAM~\cite{sam} and its efficient
variants~\cite{mobilesam,efficientsam2023,fastsam2023}
seemingly resolve this limitation by enabling category-agnostic
segmentation via spatial prompts.
However, our ablation study (Section~\ref{sec:ablation}) reveals
that na\"ively feeding a foundation segmenter with raw detection
prompts---without explicit temporal stabilization---increases
trajectory error by up to $6\times$, with the error distribution
exhibiting a pronounced long tail of catastrophic drift episodes
(Fig.~\ref{fig:violin}).
The root cause is not the segmentation model itself, but the
\emph{temporal instability of its input prompts}: under partial
occlusion and detector noise, bounding-box prompts become
intermittent or spatially jittery across frames, which propagates
through masks into inconsistent dynamic-region estimates and
ultimately destabilizes geometric estimation.
In other words, the effective bottleneck in
foundation-model-assisted dynamic VIO lies at the
\emph{perception-to-geometry interface}---specifically, the
temporal coherence of the information pathway from object
detection to geometric constraints.

This observation leads to a clear design principle: to exploit
universal segmentation for robust dynamic VIO, one must explicitly
stabilize the entire prompt--mask--geometry information chain
rather than loosely coupling off-the-shelf components.
We therefore propose \textbf{STAG-VIO} (Stabilized
Tracking-Assisted Generalized-masking VIO), which formulates
dynamic robustness as an \emph{interface stabilization} problem
and decomposes it into three tightly coupled stages, each
targeting a specific failure mode in the information pathway:
\textbf{(i)}~\emph{Track-to-Prompt} models prompt generation as
adaptive state estimation to produce temporally coherent
bounding-box prompts under partial occlusion;
\textbf{(ii)}~\emph{Prompt-to-Mask} converts stable prompts into
conservative, geometry-oriented dynamic masks via foundation
segmentation with asymmetric morphological refinement;
and \textbf{(iii)}~\emph{Mask-to-Geometry} translates refined
masks into reliable geometric constraints through budget-aware
static feature redistribution that maintains estimator
observability when dynamic regions dominate the view.
Moreover, the temporal coherence of stabilized prompts enables
a practical efficiency gain: the segmentation model's image
encoder can be invoked at reduced frequency while the lightweight
mask decoder runs every frame, as the stable prompts compensate
for slightly stale image embeddings
(Section~\ref{sec:prompt-to-mask}).
Crucially, the framework is model-agnostic: the detector and
segmentation backbone can each be replaced by alternative
implementations, provided they satisfy the interface contracts of
temporally stable prompts and geometry-reliable masks.

We validate STAG-VIO on two complementary benchmarks:
VIODE~\cite{VIODE} (outdoor, controlled dynamic levels)
and OpenLORIS-Scene~\cite{openloris} (indoor, open-world
dynamics with crowds and occlusions), demonstrating consistent
accuracy and robustness gains over state-of-the-art VIO baselines
in both settings.
Crucially, ablation reveals that the proposed prompt
stabilization is the single most impactful component, reducing
trajectory error by up to 83\%---strongly suggesting that
temporal prompt coherence, rather than segmentation model
capacity, is the dominant factor in robust dynamic VIO.
Real-world deployment on a mobile robot in complex urban traffic
further confirms the framework's practical effectiveness, with a
57.9\% RMSE reduction compared to the baseline. Our contributions are summarized as follows:
\begin{itemize}[leftmargin=1.5em, itemsep=0.2em, topsep=0.2em]
\item A perception-to-geometry interface framework is proposed
  that decomposes dynamic VIO robustness into a principled
  Track-to-Prompt $\to$ Prompt-to-Mask $\to$ Mask-to-Geometry
  pipeline, systematically stabilizing the
  detection-to-constraint pathway.
\item We introduce an uncertainty-adaptive prompt stabilization
  module that models prompt generation as state estimation with
  bounded noise adaptation, enabling temporally coherent
  segmentation under occlusion and safe encoder embedding reuse
  for improved throughput.
\item Extensive experiments across controlled, open-world, and
  real-world scenarios demonstrate consistent improvements over
  state-of-the-art baselines, with ablation revealing prompt
  coherence as the dominant accuracy factor in
  foundation-model-assisted dynamic VIO.
\end{itemize}

\section{Related Work}

Handling dynamic objects in VIO and visual SLAM has been
approached from three complementary directions:
geometry-only filtering, semantic-assisted rejection, and
joint perception--estimation frameworks.

\textbf{Geometry-based methods} identify and reject dynamic
measurements without external semantic cues. RP-VIO~\cite{rp-vio}
detects inconsistencies between visual reprojection and IMU
prediction to flag outliers.
RD-VIO~\cite{rdvio2023} proposes an IMU-PARSAC algorithm that
performs two-stage robust matching using accumulated
motion statistics, handling both dynamic scenes and pure
rotation.
DynaVINS~\cite{dyna-vins} formulates a novel loss function in
bundle adjustment to downweight features whose motion
is inconsistent with IMU preintegration, avoiding the need
for explicit object detection.
DFF-VIO~\cite{dff-vio} further introduces a dynamic feature
fusion strategy that adaptively weights features based on
motion consistency, enabling feature reuse rather than
outright rejection.
While effective in moderately dynamic scenes, these methods
rely on motion consistency as their sole discriminative
signal; when large portions of the field of view move
coherently (e.g., a bus occupying 60\% of the image), the
consensus set itself is corrupted and estimation degrades.

\textbf{Semantic-assisted methods} leverage learned
segmentation to identify known dynamic classes and mask
their features before geometric estimation.
DS-SLAM~\cite{dsslam2018} augments ORB-SLAM2 with a
SegNet~\cite{SegNet}-based semantic thread to detect
potentially dynamic objects, then applies geometric
validation to confirm their motion.
DynaSLAM~\cite{dynaslam2018} employs Mask~R-CNN~\cite{Mask_rcnn}
for multi-class instance segmentation and combines it
with multi-view geometry to handle both known and
unknown moving objects.
PVO~\cite{PVO} takes a joint approach, coupling
panoptic segmentation with visual odometry so that
segmentation benefits from geometric cues and vice versa.
These methods achieve strong results on benchmarks, yet
their segmentation networks are tied to fixed label sets
and require retraining to accommodate new environments
or object categories, limiting deployment generality.

\textbf{Foundation-model-assisted methods} have begun to
adopt promptable segmentation models to overcome the
category limitation.
SamSLAM~\cite{sam-slam} integrates SAM with epipolar-based
dynamic detection within ORB-SLAM3 to filter dynamic
features without relying on fixed label sets.
However, existing foundation-model-assisted approaches
treat prompt generation as a preprocessing step and feed
raw detection outputs directly to the segmenter, without
considering how detection noise and partial occlusion
cause prompts to jitter or vanish across frames.
As we show in Section~\ref{sec:ablation}, this
temporal prompt instability---rather than segmentation
model capacity---is the dominant factor limiting
trajectory accuracy, yet it has received no explicit
treatment in the literature.


\section{METHODOLOGY}

We present STAG-VIO, a dynamic visual--inertial odometry framework
that treats robustness in dynamic scenes as an explicit
\emph{perception-to-geometry interface} stabilization problem.
Given a synchronized image stream $\{I_t\}$ and IMU measurements
$\{u_t\}$, STAG-VIO estimates the platform state $\mathbf{x}_t$
(pose, velocity, and IMU biases) while explicitly stabilizing the
information pathway from dynamic object detection to geometric
constraints---the perception-to-geometry interface whose
instability we identified in Section~\ref{sec:introduction} as
the dominant accuracy bottleneck.

\subsection{Overview: Perception-to-Geometry Information Chain}
\label{sec:overview}
\begin{figure*}[!t]
  \centering
  \setlength{\abovecaptionskip}{0.0cm}
   \includegraphics[width=1.0   \textwidth]{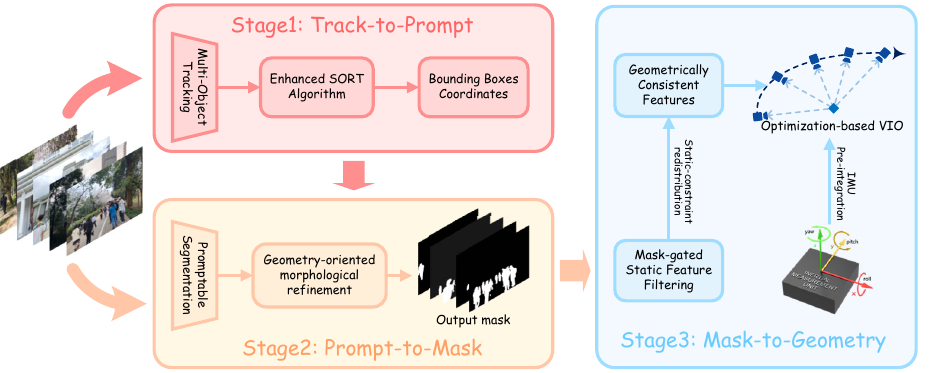}
   \caption{Overview of the STAG-VIO pipeline. The three stages
address successive instabilities in the perception-to-geometry
information chain: \textbf{Stage\,1} stabilizes raw detections
into temporally coherent box prompts via uncertainty-adaptive
tracking; \textbf{Stage\,2} converts prompts into conservative
dynamic masks through promptable segmentation and
geometry-oriented morphological refinement;
\textbf{Stage\,3} filters dynamic features and redistributes
static constraints to maintain well-conditioned
visual--inertial optimization.}
   \label{fig:pipline}
\end{figure*}

STAG-VIO decomposes dynamic robustness into three coupled stages
along the information chain
$\mathcal{D}_t \!\to\! \hat{\mathcal{B}}_t \!\to\!
\mathbf{M}_t^{\mathrm{ref}} \!\to\!
\mathcal{P}_t^{\mathrm{static}} \!\to\! \mathbf{x}_t$,
where raw detections $\mathcal{D}_t$ are converted into
stabilized prompts $\hat{\mathcal{B}}_t$, then refined masks
$\mathbf{M}_t^{\mathrm{ref}}$, and finally a static feature set
$\mathcal{P}_t^{\mathrm{static}}$ for the back-end optimizer.

A critical observation is that noise introduced at any stage
propagates and \emph{amplifies} downstream: jittery detections
produce flickering prompts, which yield temporally inconsistent
masks, which in turn inject motion-corrupted features into
geometric estimation.
Our design inserts an explicit stabilization mechanism at each
transition:
\textbf{(i)}~\emph{Track-to-Prompt} converts noisy frame-wise
detections into temporally coherent box prompts via
uncertainty-adaptive multi-object tracking;
\textbf{(ii)}~\emph{Prompt-to-Mask} obtains category-agnostic
pixel masks from stable prompts and applies geometry-oriented
morphological refinement to establish conservative safety margins;
\textbf{(iii)}~\emph{Mask-to-Geometry} translates refined masks
into reliable geometric constraints through mask-gated feature
filtering and a constraint-budget-aware redistribution strategy.

This decomposition explicitly separates open-world perception
from metric estimation while enforcing a stable interface
between them.

\subsection{Track-to-Prompt: Occlusion-Robust Prompt
            Stabilization via Uncertainty-Adaptive Tracking}
\label{sec:track-to-prompt}

Let $\mathcal{B}_t = \{\mathbf{z}_t^k\}$ denote the set of
candidate bounding boxes at time~$t$ produced by a real-time
object detector.
As discussed in Section~\ref{sec:introduction}, directly using
$\mathcal{B}_t$ as prompts is brittle: detections become
intermittent or jittery under occlusion, producing flickering
masks that destabilize geometric estimation.
However, the physical motion of tracked objects is inherently
smooth, suggesting that prompt generation can be modeled as a
\emph{state estimation} problem where a predictive filter bridges
detection gaps and suppresses jitter, yielding stabilized prompts
$\hat{\mathcal{B}}_t$.

\subsubsection{State model}
For each tracked object~$k$, we maintain a state vector
\begin{equation}
\label{eq:state}
\mathbf{x}^k_t =
  \begin{bmatrix}
    x, & y, & w, & h, & v_x, & v_y, & v_w, & v_h
  \end{bmatrix}^\top,
\end{equation}
where $(x,y)$ is the box center, $(w,h)$ are the box dimensions,
and $(v_x, v_y, v_w, v_h)$ are their respective temporal rates.
A constant-velocity Kalman filter predicts
$\mathbf{x}^k_{t|t-1}$ and updates it upon association with a
matched detection.

\subsubsection{Data association}
We compute inter-frame correspondences between detections and
predicted tracks by solving a global assignment problem via
Hungarian matching~\cite{rajabi2023computing} under a geometric IoU
cost. This yields a consistent correspondence map that supports
multi-object prompt continuity even when individual detections
are momentarily missed.

\subsubsection{Uncertainty-adaptive prompt stabilization}

In practice, detection reliability varies significantly across
frames.
During partial occlusion, the detector may report a
shrunken or shifted box; under severe occlusion, the detection
may vanish entirely. If the Kalman filter's measurement noise
covariance $\mathbf{R}$ is fixed, it either over-trusts unreliable
detections (causing the prompt to jump) or under-trusts reliable
ones (causing the prompt to lag).
We therefore adaptively adjust $\mathbf{R}$ based on recent
prediction residuals, implementing an asymmetric trust mechanism:
\emph{when detections are consistent with predictions, trust them;
when they deviate substantially, downweight them}.

Specifically, for each measurement dimension~$j$ we compute the
RMSE of the innovations within a sliding window of $N$~frames:
\begin{equation}
\label{eq:rmse}
r^{(j)}_{\mathrm{RMSE}} =
  \sqrt{
    \frac{1}{N}
    \sum_{i=k-N+1}^{k}
    \bigl(\nu^{(j)}_i\bigr)^2
  }\,,
\end{equation}
where $\boldsymbol{\nu}_k = \mathbf{z}_k - \mathbf{H}\,
\hat{\mathbf{x}}_{k|k-1}$ is the innovation (prediction residual)
at time step~$k$.
The residual magnitude is then mapped to a measurement noise
value through a bounded, monotonically non-decreasing activation
function $\varphi: [0,+\infty) \to [R_{\min},\, R_{\max}]$
with $R_{\min} > 0$:
\begin{equation}
\label{eq:R_adapt}
\mathbf{R}_k = \beta \cdot \mathrm{diag}\!\Bigl(
  \varphi\bigl(\lambda \cdot r^{(1)}_{\mathrm{RMSE}}\bigr),
  \;\ldots,\;
  \varphi\bigl(\lambda \cdot r^{(m)}_{\mathrm{RMSE}}\bigr)
\Bigr),
\end{equation}
where $\beta$ controls the global noise magnitude and $\lambda$
governs sensitivity to residual changes.

\textbf{Design rationale for the bounds.}\;
The positive lower bound $R_{\min}$ prevents the Kalman gain from
diverging when residuals are momentarily small---for instance,
immediately after an occlusion ends, the first matched detection
may coincidentally align with the prediction, and an unbounded
gain would cause the filter to over-commit to this single
observation, amplifying prompt jitter on subsequent frames.
Conversely, the upper bound $R_{\max}$ ensures that observations
are never entirely discarded: even under severe, prolonged
occlusion, the track retains a nonzero update from any matched
detection, allowing prompt recovery once the target reappears.

\textbf{Choice of activation function.}\;
We adopt
$\varphi(x) = R_{\min} + (R_{\max} - R_{\min})\,\mathrm{erf}(x)$,
where $\mathrm{erf}(\cdot)$ is the Gauss error function.
Unlike arbitrary logistic curves (\emph{e.g.}, sigmoid or tanh),
the error function naturally aligns with the Gaussian noise
assumption underlying the Kalman filter: it represents the
cumulative probability of the measurement error magnitude
exceeding nominal bounds, providing a principled, rather than
\emph{ad hoc}, mapping from residual statistics to noise
covariance.
In practice, the overall filtering behavior is primarily governed
by $\beta$, $\lambda$, $R_{\min}$, and $R_{\max}$; replacing
$\mathrm{erf}$ with other saturating functions yields comparable
results, confirming that the bounded adaptation mechanism matters
more than the specific functional form.

\subsection{Prompt-to-Mask: Category-Agnostic Dynamic Parsing
            with Geometry-Oriented Refinement}
\label{sec:prompt-to-mask}

Given stabilized prompts $\hat{\mathcal{B}}_t$, we obtain
pixel-level dynamic masks using a promptable foundation
segmentation model with box prompting~\cite{mobilesam, mobilesamv2}.
For each prompt $\hat{\mathbf{b}}^k_t$, the model predicts an
instance mask
$\mathbf{m}^k_t \in \{0,1\}^{H \times W}$.
The overall dynamic mask is the union across all tracked instances:
\begin{equation}
\label{eq:union_mask}
\mathbf{M}_t = \bigvee_k \mathbf{m}^k_t\,.
\end{equation}

\subsubsection{Why recognition-optimal masks hurt geometry}

Promptable segmentation models are trained to produce
pixel-accurate, tightly fitting masks---optimal for recognition
but problematic for geometry.
Features near the contour of a moving object are often corrupted
by motion blur, cast shadows, or minor boundary inaccuracies;
if misclassified as static, they inject motion-corrupted
measurements into the optimizer.
Crucially, the cost is \emph{strongly asymmetric}: a false
negative (leaking a dynamic pixel) directly introduces a
corrupted residual that can trigger catastrophic drift, whereas
a false positive (over-masking a static pixel) merely reduces
the constraint budget---a loss explicitly recoverable by
redistribution (Section~\ref{sec:mask-to-geometry}).
This asymmetry motivates a deliberately conservative masking
strategy.

\subsubsection{Geometry-oriented morphological refinement}

To operationalize the conservative masking principle, we refine
the raw union mask $\mathbf{M}_t$ using an asymmetric
morphological strategy.
Instead of the conventional denoise-then-expand pipeline, we
\emph{first} apply an aggressive dilation using a large
elliptical structuring element
$\mathcal{S}_d$ to explicitly inflate the dynamic boundaries, establishing
a safety buffer around every moving object:
\begin{equation}
\label{eq:dilate}
\mathbf{M}_t^{\mathrm{dilate}} = \mathbf{M}_t \oplus \mathcal{S}_d\,.
\end{equation}
This is followed by a lightweight erosion with a much smaller
element $\mathcal{S}_e$  to
smooth out boundary artifacts and jagged edges introduced by the
large-kernel dilation:
\begin{equation}
\label{eq:refine}
\mathbf{M}_t^{\mathrm{ref}} =
  \mathbf{M}_t^{\mathrm{dilate}} \ominus \mathcal{S}_e\,.
\end{equation}

By design, $\mathcal{S}_d \gg \mathcal{S}_e$, so this
``heavy-dilate, light-erode'' operation produces masks that
\emph{over-envelope} the moving targets, providing a robust
geometry-oriented buffer that actively prevents
motion-corrupted edge features from leaking into the
visual--inertial back-end.
The static features lost to this conservative expansion are
explicitly recovered by the constraint-preserving redistribution
strategy in the next stage, ensuring that the overall constraint
budget is maintained.

\subsubsection{Embedding reuse via prompt-driven scheduling}

The image encoder of the promptable segmentation model
dominates per-frame computation, whereas the
prompt-conditioned mask decoder is lightweight.
Because the Track-to-Prompt stage produces temporally
coherent prompts, the visual context relevant to dynamic
objects evolves smoothly across adjacent frames, allowing
the image embedding to be safely cached and reused.
We therefore invoke the encoder only once every $K$~frames;
in intermediate frames, the mask decoder runs against the
cached embedding with current stabilized prompts.

This decoupling relies on the mask decoder receiving two
inputs: the image embedding and the box prompt.
When the embedding is slightly stale, the up-to-date
stabilized prompt still anchors the decoder to the correct
region---without prompt stabilization, both inputs would
be unreliable simultaneously, making reuse infeasible.
The tolerable staleness depends on scene appearance:
well-lit conditions permit longer reuse, whereas
low-contrast scenes narrow the safe window.
We empirically determine $K$ in
Section~\ref{sec:runtime}.

\subsection{Mask-to-Geometry: Constraint-Preserving Feature
            Processing under Large Dynamic Regions}
\label{sec:mask-to-geometry}

The refined dynamic mask $\mathbf{M}_t^{\mathrm{ref}}$ is
injected into the VIO front-end to gate visual measurements:
tracked features propagated to frame~$t$ by sparse optical flow
are discarded whenever they fall inside $\mathbf{M}_t^{\mathrm{ref}}$,
preventing motion-corrupted residuals from entering the
estimator.

However, this masking introduces a \emph{second-order failure
mode}.
When dynamic regions occupy a large portion of the image---a
common situation in crowded urban or indoor scenes---naive masking
decimates the tracked feature set.
The surviving static features tend to cluster in narrow image
bands, degrading the spatial conditioning of the visual--inertial
optimization and, in extreme cases, rendering it ill-posed.
Existing approaches either accept this constraint loss (risking
estimator divergence) or relax the mask to re-admit potentially
corrupted measurements.
We instead adopt a third strategy: \emph{actively managing the
constraint budget} by replenishing and redistributing static
features within the unmasked region.

\subsubsection{Constraint-preserving redistribution}

To sustain a robust geometric constraint budget, we replenish
the tracked feature set to a target capacity $N_{\max}$ within
the static region
$\overline{\mathbf{M}_t^{\mathrm{ref}}}$, excluding existing
tracked points.
Because the remaining static area is often narrow, newly
detected features tend to cluster, degrading geometric
conditioning.
We therefore extract surplus ORB candidates~\cite{orb}
and apply Adaptive Non-Maximal Suppression
(ANMS)~\cite{ANMS}, which selects keypoints by their minimum
suppression radius
$r_i = \min_{j:\, R_j > R_i} \|\mathbf{p}_i - \mathbf{p}_j\|_2$
to enforce approximately uniform spatial spread.
A final distance filter $D_{\min}$ removes remaining close
pairs, ensuring a well-conditioned static constraint set even
when the dynamic mask covers the majority of the image.

\subsection{Back-End Visual--Inertial Estimation}
\label{sec:backend}

The back-end follows a standard optimization-based VIO formulation
with IMU pre-integration and reprojection residuals, but it
operates on the \emph{mask-filtered and redistributed} static
feature set:
\begin{equation}
\label{eq:static_set}
\mathcal{P}_t^{\mathrm{static}} =
  \bigl\{
    \mathbf{p} \in \mathcal{P}_t
    \;\big|\;
    \mathbf{M}_t^{\mathrm{ref}}(\mathbf{p}) = 0
  \bigr\}\,.
\end{equation}
By suppressing dynamic outliers at the \emph{interface level}
rather than relying on the optimizer's internal robustification,
the back-end receives a stable stream of reliable geometric
constraints even under severe occlusion and diverse moving objects.

\textbf{Model agnosticism.}\;
Our framework imposes no assumptions on the specific detector or segmentation model, as long as they collectively
produce temporally stable prompts and geometry-reliable masks.
The detector can be replaced by any real-time proposal generator
(\emph{e.g.}, YOLO series~\cite{yolov5v8v10,yolov11},
RT-DETR series~\cite{RT-DETR,RT-DETRv2}); and the segmenter by other promptable foundation models
(\emph{e.g.}, EfficientSAM~\cite{efficientsam2023},
FastSAM~\cite{fastsam2023},
SAM~2~\cite{sam2}).
This modularity enables flexible trade-offs between accuracy and
computational efficiency depending on deployment constraints.


\section{EXPERIMENTS}
\label{sec:experiments}

\subsection{Experimental Protocol}
\label{sec:protocol}

\textbf{Datasets.}\;
We evaluate STAG-VIO in both outdoor and indoor dynamic
settings.

\textbf{(1)~VIODE}~\cite{VIODE} provides controlled
dynamic levels (\textit{none / low / mid / high}) across
urban environments, enabling systematic stress tests as
moving objects increasingly dominate the field of view.
We report results on three environments---\textit{City~day},
\textit{City~night}, and \textit{Parking~lot}
(12~sequences in total)---which together cover the key
challenges of high dynamic density, significant illumination
variation, and diverse scene geometry.

\textbf{(2)~OpenLORIS-Scene}~\cite{openloris} contains
long-horizon indoor sequences across five diverse settings
(\textit{office, corridor, home, cafe, market}) with realistic
challenges including lighting variations, crowds, and frequent
partial occlusions.

\textbf{Metrics.}\;
On VIODE, we report \textbf{RMSE of Absolute Trajectory Error
(ATE)} in meters.
On OpenLORIS-Scene, we additionally report \textbf{Correct Rate
(CR)}~\cite{openloris} over the full time span, where CR
reflects tracking robustness (higher is better) and ATE~RMSE
reflects accuracy (lower is better).

\textbf{Implementation details.}\;
All experiments are conducted on an NVIDIA RTX~4090D GPU. We instantiate STAG-VIO with
YOLOv11x~\cite{yolov11} for object detection and
MobileSAM~\cite{mobilesam} for prompt-guided
segmentation.
For the uncertainty-adaptive tracker, the sliding window size is $N=10$,
sensitivity $\lambda=0.01$, and noise scale $\beta=1.0$; the bounded activation
function $\varphi$ is instantiated with the Gauss error function as in
Eq.~\eqref{eq:R_adapt}.
For geometry-oriented morphological refinement, the dilation
kernel $\mathcal{S}_d$ is a $25\!\times\!25$ elliptical element
and the erosion kernel $\mathcal{S}_e$ is $3\!\times\!3$.
Feature parameters are set to $N_{\max}\!=\!150$ and
$D_{\min}\!=\!15$\,pixels unless otherwise stated.
The image encoder is invoked every $K\!=\!2$ frames by default
.

\subsection{Main Results on Public Benchmarks}
\label{sec:main_results}

\begin{table*}[t]
    \caption{Comparison with State-of-the-art Methods (RMSE of ATE in [M]). 
    We highlight the best of each column in \textcolor{bestred}{red}. 
    The STAG-VIO row is highlighted in light yellow, and the three \textit{high} columns are highlighted in light blue.}
    \centering
    \resizebox{\textwidth}{!}{%
    \begin{tabular}{c|
    >{\centering\arraybackslash}p{3.0em}
    >{\centering\arraybackslash}p{3.0em}
    >{\centering\arraybackslash}p{3.0em}
    >{\columncolor{highcol}\centering\arraybackslash}p{3.0em}|
    >{\centering\arraybackslash}p{3.0em}
    >{\centering\arraybackslash}p{3.0em}
    >{\centering\arraybackslash}p{3.0em}
    >{\columncolor{highcol}\centering\arraybackslash}p{3.0em}|
    >{\centering\arraybackslash}p{3.0em}
    >{\centering\arraybackslash}p{3.0em}
    >{\centering\arraybackslash}p{3.0em}
    >{\columncolor{highcol}\centering\arraybackslash}p{3.0em}}
    \toprule
    \multicolumn{1}{c|}{} & \multicolumn{12}{c}{\textbf{VIODE}} \\
    \cmidrule{2-13}
    \multicolumn{1}{c|}{\textbf{Method}} & \multicolumn{4}{c|}{\textbf{City day}} & \multicolumn{4}{c|}{\textbf{City night}} & \multicolumn{4}{c}{\textbf{Parking lot}} \\
    \cmidrule{2-13}
    \multicolumn{1}{c|}{} & \textbf{none} & \textbf{low} & \textbf{mid} & \textbf{high} 
    & \textbf{none} & \textbf{low} & \textbf{mid} & \textbf{high} 
    & \textbf{none} & \textbf{low} & \textbf{mid} & \textbf{high} \\
    \midrule
    ORB-SLAM3~\cite{orb-slam3}
      & 2.179 & 5.301 & 2.204 & *
      & 0.176 & * & * & *
      & 0.287 & 2.897 & 6.742 & 7.482 \\
    VINS-Fusion~\cite{vins-fusion}
      & 0.203 & \cellcolor{bestred} 0.148 & 0.261 & 0.321
      & 0.319 & 0.368 & 0.433 & 0.490
      & 0.120 & 0.113 & \cellcolor{bestred} 0.161 & 1.201 \\
    VINS-Mono~\cite{vins-mono}
      & 0.186 & 0.237 & 0.263 & 3.169
      & 0.314 & 0.436 & 0.727 & 0.575
      & 0.102 & \cellcolor{bestred} 0.109 & 2.915 & 4.933 \\
    VINS-Mah~\cite{vins-mah}
      & 0.149 & 0.193 & 0.222 & 0.264
      & 0.417 & 0.610 & 0.651 & 0.556
      & 0.136 & 0.151 & 0.246 & 0.128 \\
    RP-VIO~\cite{rp-vio}
      & 0.382 & 0.229 & 0.435 & 0.536
      & 0.263 & 0.509 & 0.652 & 0.577
      & 0.981 & 1.334 & 0.375 & 0.713 \\
    RD-VIO~\cite{rdvio2023}
      & 0.305 & \cellcolor{bestred} 0.148 & 0.187 & 1.221
      & 0.305 & 0.871 & 0.635 & 0.560
      & 0.299 & 0.359 & 0.312 & 0.527 \\
    Dyna-VINS~\cite{dyna-vins}
      & 0.349 & 0.330 & 0.258 & 0.245
      & 0.687 & \cellcolor{bestred} 0.207 & 0.251 & 0.311
      & 0.106 & 0.167 & 0.181 & 0.105 \\
    \textbf{STAG-VIO (Ours)}
      & \best{0.107} & \textbf{0.156} & \best{0.138} & \best{0.204}
      & \best{0.125} & \textbf{0.246} & \best{0.222} & \best{0.210}
      & \best{0.099} & \textbf{0.160} & \textbf{0.204} & \best{0.094} \\
    \specialrule{1.2pt}{0pt}{0pt}
    \arrayrulecolor{black}
    \multicolumn{13}{l}{\small $\bullet$ *: Failure case.}\\
    \end{tabular}%
    }%
    \label{tab:viode}%
\end{table*}

\begin{figure*}[!t]
  \centering
  \subfloat[City day]
  {
        \label{fig:traj(a)}
        \includegraphics[width=0.3\textwidth]{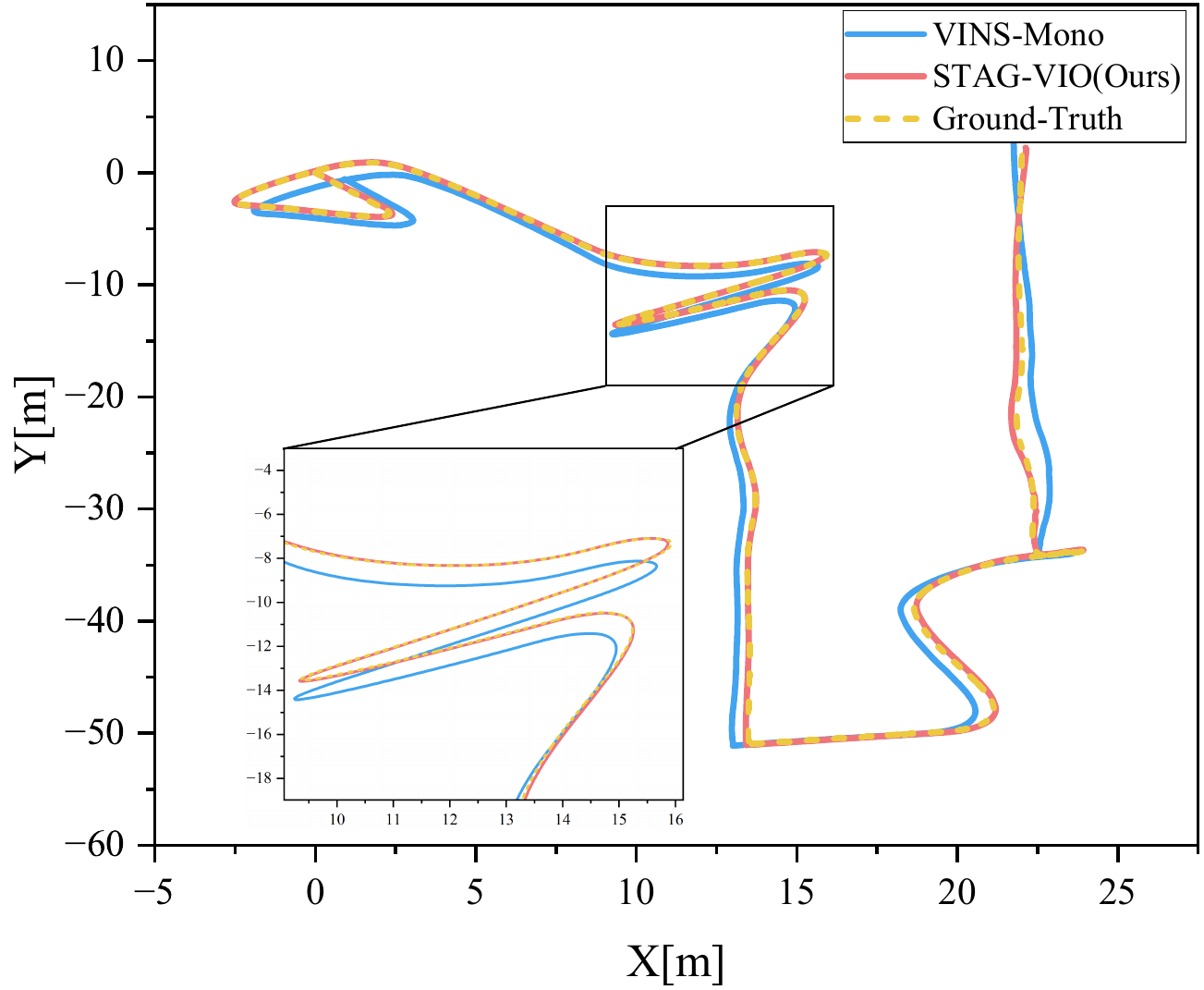}
  }
  \subfloat[City night]
  {
        \label{fig:traj(b)}
        \includegraphics[width=0.3\textwidth]{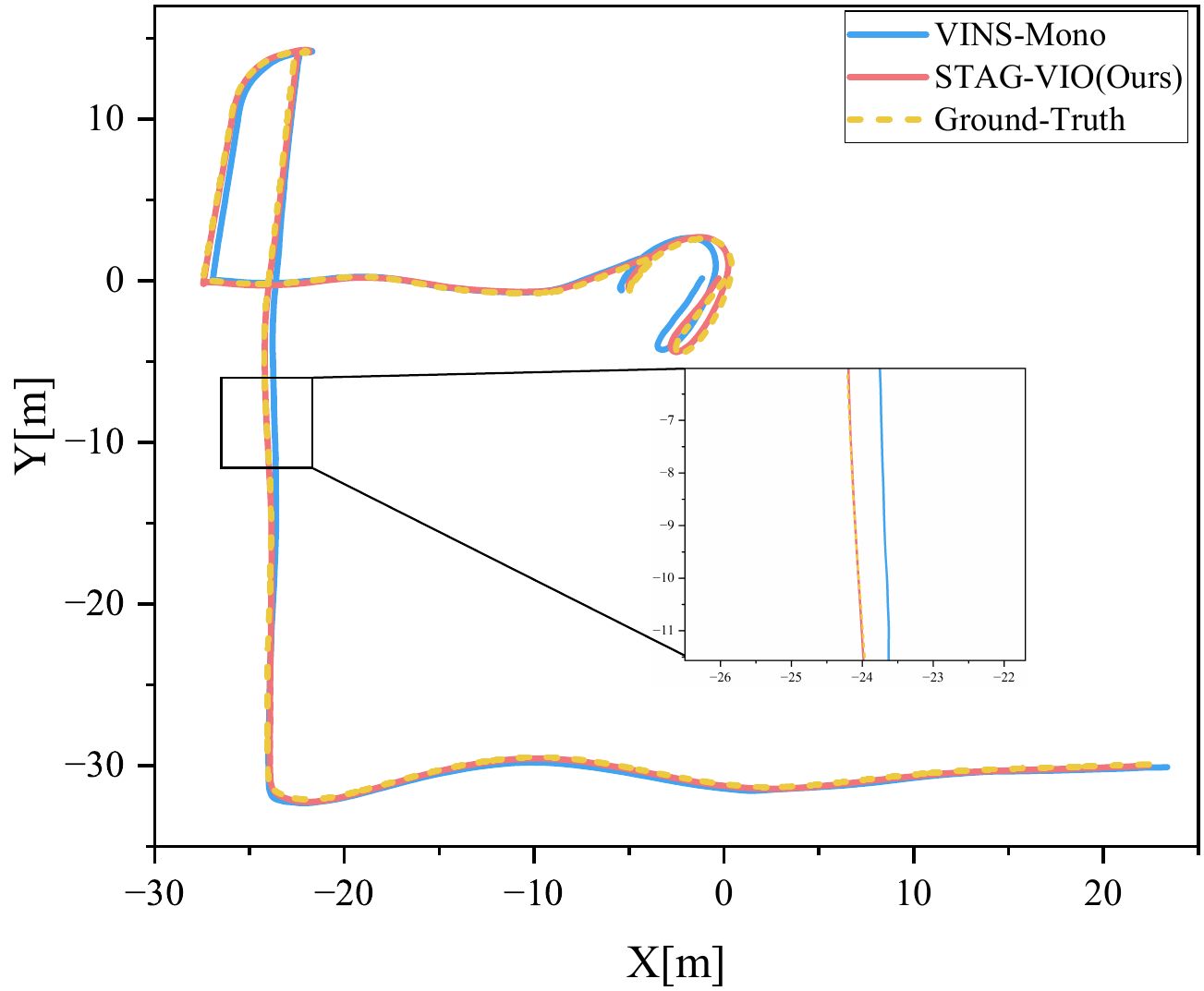}
  }
  \subfloat[Parking lot]
  {
        \label{fig:traj(c)}
        \includegraphics[width=0.3\textwidth]{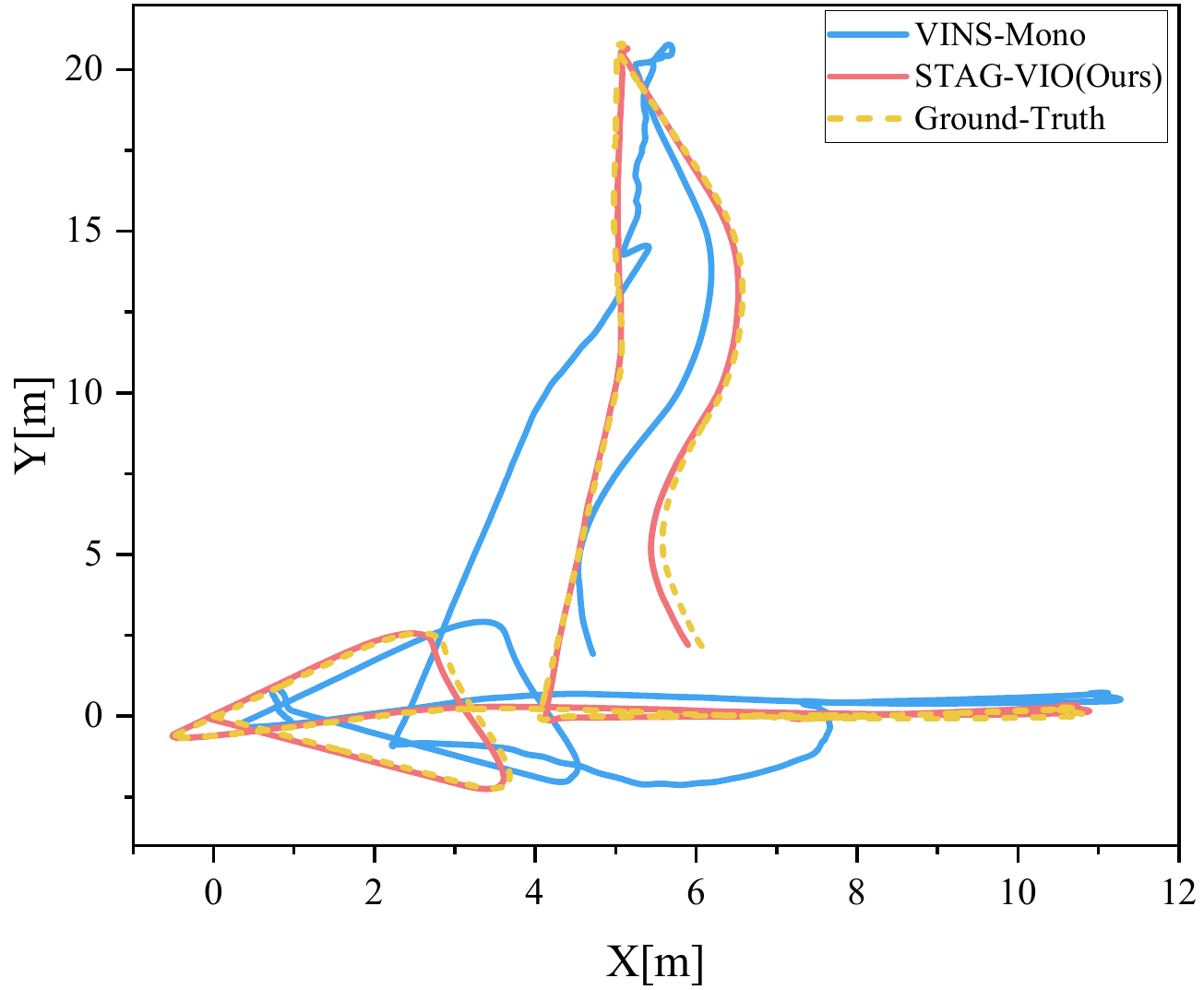}
  }
\caption{Trajectory comparison on the high-dynamic VIODE sequences, including City day, 
City night, and Parking lot. The ground truth is shown as a dashed yellow line. 
STAG-VIO closely follows the ground-truth trajectories across all scenes, while 
VINS-Mono exhibits significant drift under strong dynamic interference.}
  \label{fig:traj}
\end{figure*}

\subsubsection{VIODE: accuracy under controlled dynamics}

Table~\ref{tab:viode} reports ATE RMSE across all four dynamic levels
(none/low/mid/high) for the three VIODE environments. STAG-VIO achieves the
best result in 8 of 12 settings and, more importantly, never fails and never
exceeds 0.246\,m ATE in any configuration---including the most adversarial
high-dynamic sequences---whereas every baseline exhibits at least one
catastrophic failure or divergence.

\textbf{Graceful degradation vs.\ catastrophic collapse.} The defining
difference between STAG-VIO and the baselines is not average accuracy but the
\emph{shape} of degradation as dynamics intensify. Purely geometric or
fixed-vocabulary methods stay competitive up to a threshold and then collapse
abruptly: VINS-Mono is accurate on Parking lot at none/low (0.102/0.109\,m) but
jumps to 2.915\,m at mid and 4.933\,m at high---a $45\times$ increase---while
ORB-SLAM3 fails outright ($*$) on City day--high and on all dynamic City night
sequences. RD-VIO (0.187$\rightarrow$1.221\,m from mid to high on City day) and
VINS-Fusion (0.161$\rightarrow$1.201\,m on Parking lot) show the same cliff
behavior. In contrast, STAG-VIO degrades gracefully: across the full
none$\rightarrow$high sweep its error stays within 0.094--0.246\,m in every
environment, a spread of under $2.2\times$, indicating that the
perception-to-geometry interface removes the dominant failure mode rather than
merely lowering the mean.

\textbf{High-dynamic regime.} The advantage is most pronounced under high
dynamics, where moving objects dominate the field of view. Using
VINS-Mono~\cite{vins-mono} as a consistent reference, STAG-VIO reduces ATE RMSE
from 3.169\,m to 0.204\,m on City day--high (93.56\%), from 0.575\,m to
0.210\,m on City night--high (63.48\%), and from 4.933\,m to 0.094\,m on
Parking lot--high (98.09\%). These gains are not merely relative to a weak
baseline: even against the \emph{strongest} competing method in each
high-dynamic column (Dyna-VINS in all three), STAG-VIO still improves accuracy
by 16.7\% (City day), 32.5\% (City night), and 10.5\% (Parking lot), confirming
that the benefit stems from the stabilized interface rather than from an
unfavorable baseline choice.

\textbf{The low-dynamic regime.} In the four settings where STAG-VIO does not
rank first---City day--low, City night--low, Parking lot--low, and Parking
lot--mid---the scenes are predominantly low-dynamic, where dynamic masking
offers little benefit and the modest overhead of the segmentation pipeline can
marginally affect estimation. Crucially, the absolute gaps are
small in every case ($\leq$0.051\,m; e.g., 0.156 vs.\ 0.148\,m on City
day--low), and the methods that edge out STAG-VIO in these near-static
conditions are precisely those that collapse once dynamics increase (VINS-Mono
wins Parking lot--low at 0.109\,m but diverges to 4.933\,m at high). STAG-VIO
therefore trades a negligible, bounded overhead in easy scenes for a decisive
robustness advantage across the operating envelope---a favorable trade-off for
deployment, where worst-case rather than best-case behavior governs safety.
Figure~\ref{fig:traj} visualizes estimated trajectories on the high
sequences, where STAG-VIO consistently aligns with the ground truth while
VINS-Mono exhibits severe drift.

\subsubsection{OpenLORIS-Scene: long-horizon robustness in open-world indoor scenes}

\begin{figure*}[!t]
  \centering
  \setlength{\abovecaptionskip}{0.0cm}
   \includegraphics[width=0.95\textwidth]{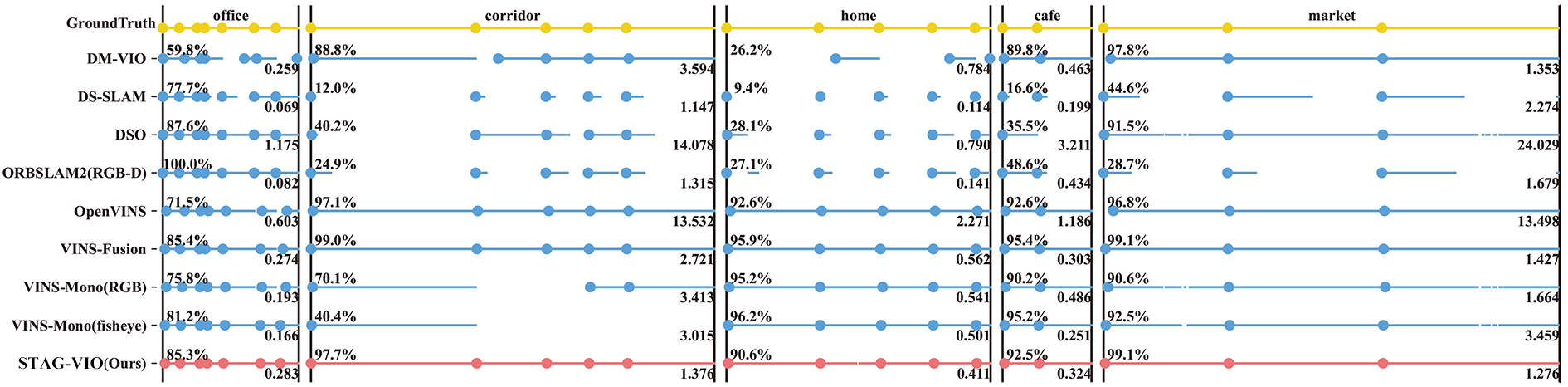}
   \caption{Experiment results with the OpenLORIS-Scene datasets. For every algorithm, the dots represent successful initiation instances, and the lines show the span of successful tracking. The percentage in the top left corner of each scene indicates the average correct rate; a higher correct rate suggests greater algorithm robustness. The float value on the first line below depicts the average ATE RMSE, with lower values indicating higher accuracy. Some experimental results from prior methods are referenced from ~\cite{openloris}}
   \label{fig:openloris}
\end{figure*}

Figure~\ref{fig:openloris} reports both Correct Rate (CR) and ATE RMSE across
five indoor scenes. Unlike the controlled VIODE benchmark, OpenLORIS-Scene
consists of long-horizon sequences recorded in everyday environments, where the
dominant challenges are sustained tracking over extended durations, open-world
object categories, and frequent partial occlusions rather than a single
controllable dynamic level. We therefore read the two metrics jointly: CR
measures how large a fraction of the trajectory is tracked without losing
initialization (robustness, higher is better), while ATE RMSE measures accuracy
over the tracked portion (lower is better).

\textbf{Robustness under heavy dynamics.} The two most demanding scenes are
\emph{market} and \emph{corridor}. In \emph{market}, dominated by dense crowds
and diverse moving objects, STAG-VIO sustains 99.1\% CR with 1.276\,m ATE,
tracking almost the entire sequence despite pervasive dynamic occlusion---the
regime our perception-to-geometry interface is designed for, and where
fixed-vocabulary baselines tend to lose tracking as unseen or heavily occluded
objects evade their segmenters. In \emph{corridor}, characterized by low
texture, illumination changes, and frequent occlusions, STAG-VIO maintains
97.7\% CR with 1.376\,m ATE, showing resilience to the compounded stress of
weak visual constraints and dynamic interference.

\textbf{Balanced performance across scenes.} We note that the highest-CR scenes
(\emph{market}, \emph{corridor}) also show larger absolute ATE than shorter,
more static scenes such as \emph{office}; this is expected rather than
contradictory, as sustaining tracking through long, heavily dynamic
trajectories accumulates more drift than terminating early would, and high CR
paired with meter-level drift is the more deployment-relevant outcome. Across
the remaining scenes (\emph{office}, \emph{home}, \emph{cafe}), STAG-VIO
preserves a favorable robustness--accuracy balance, sustaining high CR while
keeping ATE competitive with state-of-the-art methods. Taken with the VIODE
results, this consistency across controlled outdoor dynamics and uncontrolled
open-world indoor conditions indicates that stabilizing the
perception-to-geometry interface generalizes beyond any single benchmark.

\subsection{Analysis of the Perception-to-Geometry Interface}
\label{sec:ablation}
\subsubsection{Impact of prompt stabilization}

\begin{figure}[!tbp]
    \centering
        \includegraphics[width=1.0\linewidth]{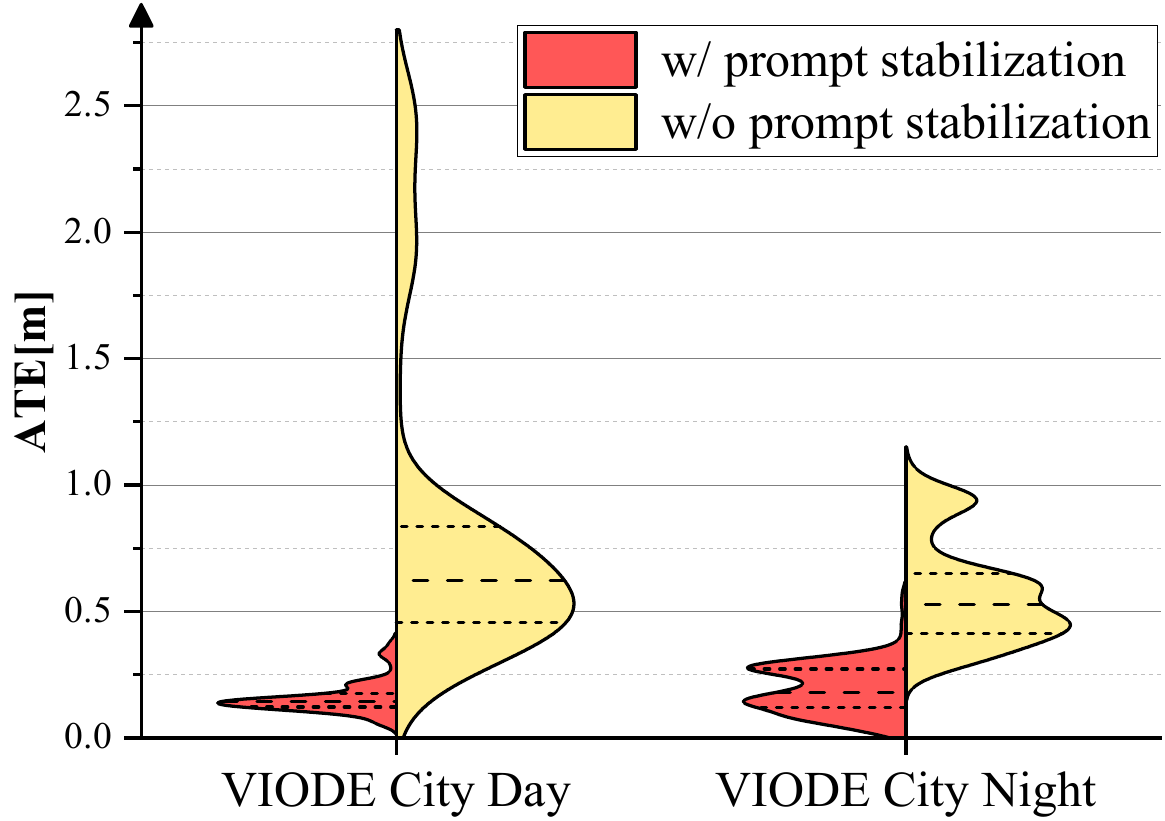}
        \caption{
        Violin plots of ATE on VIODE (City day/night) comparing STAG-VIO with and without the proposed prompt-stabilization tracker, showing that stabilized prompts yield lower and more consistent ATE.
        }\label{fig:violin}
\end{figure}

To isolate the contribution of the Track-to-Prompt stage, we ablate the
proposed prompt-stabilization tracker---feeding raw per-frame detections
directly to the segmenter---and evaluate on VIODE. Figure~\ref{fig:violin}
compares the per-frame ATE distributions with and without stabilization on City
Day and City Night; full distributional statistics are reported in
Appendix~\ref{app:ablation}.

\textbf{Accuracy gain.} Prompt stabilization reduces the RMSE of the ATE
sequence by 83.28\% on City Day and 63.99\% on City Night. The improvement is
equally evident in central tendency: the median ATE drops from 0.622\,m to
0.145\,m (76.7\%) on City Day and from 0.528\,m to 0.180\,m (65.8\%) on City
Night, confirming that the gain reflects a systematic shift of the entire
distribution rather than a change in a single summary statistic.

\textbf{Elimination of catastrophic drift.} Beyond the shift in central
tendency, the shape of the distribution changes qualitatively---the property
most visible in the violin plots. On City Day, the long upper tail collapses:
the 95th-percentile ATE falls from 2.373\,m to 0.326\,m and the maximum from
2.608\,m to 0.406\,m, while the fraction of frames exceeding 1\,m drops from
16.6\% to 0\%. Concentration tightens accordingly, with the interquartile range
shrinking $7.0\times$ (0.380\,m $\rightarrow$ 0.054\,m) and the standard
deviation $9.5\times$ (0.609\,m $\rightarrow$ 0.064\,m). City Night shows the
same trend more moderately (IQR $1.6\times$, std $2.0\times$ tighter), because
its baseline distribution lacks the heavy day-time tail to begin with. This
confirms that prompt stabilization does not merely lower average error but
eliminates the catastrophic drift episodes that populate the upper tail---a
property critical for deployment, where occasional large excursions are more
dangerous than uniformly moderate error.

Taken together, these results establish prompt temporal coherence as the single
largest contributor to STAG-VIO's accuracy, consistent with our central premise
that stabilizing the perception-to-geometry interface---rather than increasing
segmentation-model capacity---is the key to robust dynamic VIO.

\subsubsection{Robustness to the static-constraint budget}
\label{sec:robust}

\begin{figure}[!t]
    \centering
    \subfloat{
        \includegraphics[width=1.0\linewidth]{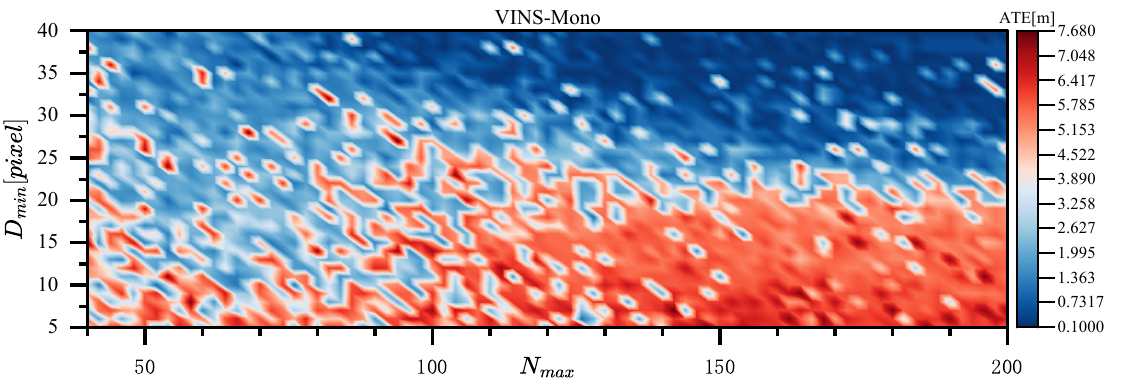}
        \label{fig5:Baseline}
    }
    \\
    \subfloat{
        \includegraphics[width=1.0\linewidth]{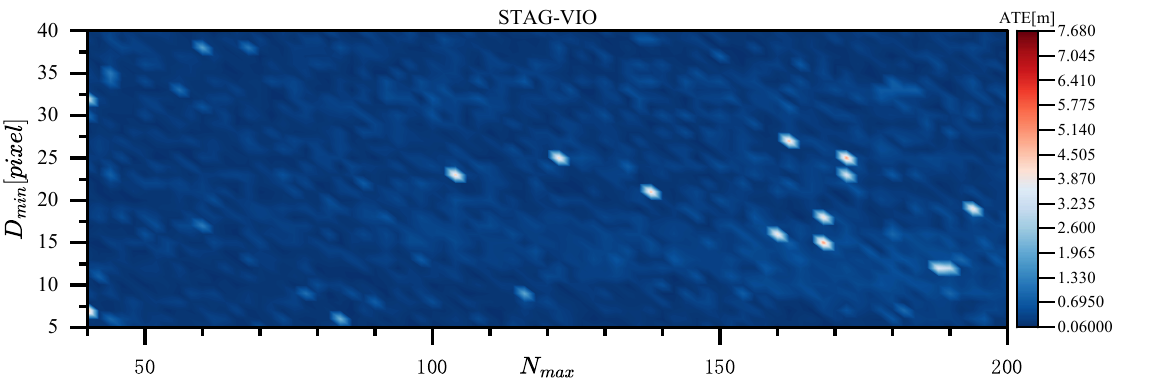}
        \label{fig5:STAG-VIO}
    }

    \caption{The heatmap illustrates RMSE ATE in relation to the maximum feature points $N_{max}$ and minimum feature point distance $D_{min}$ for the "high" sequence of the VIODE dataset. The color bar represents the range of ATE values. }
    \label{fig:heatmap}
\end{figure}

A key practical failure mode in dynamic VIO is that aggressive dynamic masking
leaves too few or poorly distributed static features. To stress-test constraint
sensitivity, we densely sweep the maximum feature count $N_{\max}\in[40,200]$
and the minimum feature distance $D_{\min}\in[5,40]$\,px, yielding 2{,}916
parameter configurations per method, and evaluate ATE RMSE on the VIODE
high-dynamic sequence. Figure~\ref{fig:heatmap} visualizes the resulting error
landscapes, and full statistics are reported in Appendix~\ref{app:robust}.

\textbf{Sensitivity of the baseline.} VINS-Mono is highly sensitive to the
constraint budget. Its RMSE spans 0.113--7.668\,m across the grid---over an
order of magnitude---with a median of 2.334\,m and an interquartile range (IQR)
of 4.127\,m. Critically, 78.9\% of all configurations exceed 1\,m error and
only 13.6\% stay below 0.5\,m, indicating that acceptable accuracy occupies a
narrow region of the parameter space. The error correlates strongly and
negatively with $D_{\min}$ ($\rho=-0.77$): without a mechanism to enforce
spatial spread, the baseline depends critically on a well-chosen minimum
distance to avoid feature clustering, and mis-setting it triggers sharp
degradation.

\textbf{Insensitivity of STAG-VIO.} In contrast, STAG-VIO remains consistently
accurate across the same space, with a median RMSE of 0.229\,m
($10.2\times$ lower than VINS-Mono) and an IQR of just 0.165\,m---a typical
spread $25\times$ tighter than the baseline. 95.2\% of all configurations stay
below 0.5\,m and only 0.8\% exceed 1\,m, so nearly the entire parameter space
is deployable rather than a narrow sweet spot. Its error is essentially
uncorrelated with both $N_{\max}$ ($\rho=0.11$) and $D_{\min}$ ($\rho=-0.06$),
showing that the mask-aware redistribution strategy does not merely remove
outliers but actively sustains a sufficient and well-distributed static
constraint set regardless of how the budget is parameterized. We attribute this
flatness to ANMS-based replenishment, which enforces spatial coverage
internally and thereby removes the baseline's dependence on $D_{\min}$.

This robustness has direct practical value, substantially reducing the need for
careful per-scenario parameter tuning. For completeness, a small fraction of
STAG-VIO configurations (0.8\%, concentrated at the largest $N_{\max}$) still
produce elevated error, where over-replenishment re-admits weakly conditioned
features; this residual sensitivity is far milder than the baseline's and is
easily avoided by keeping $N_{\max}$ in a moderate range.

\subsection{Runtime and Efficiency Analysis}
\label{sec:runtime}

\begin{table}[t]
\centering
\caption{Effect of SAM encoder interval $K$ on accuracy and throughput
(VIODE \textit{high} dynamic level, RTX~4090D).}
\label{tab:runtime}
\setlength{\tabcolsep}{6pt}
\begin{tabular}{c cc cc}
\toprule
\multirow{2}{*}{$K$}
  & \multicolumn{2}{c}{ATE RMSE [m] $\downarrow$}
  & \multicolumn{2}{c}{FPS $\uparrow$} \\
\cmidrule(lr){2-3} \cmidrule(lr){4-5}
  & City day & City night
  & City day & City night \\
\midrule
1 (base)   & 0.197          & 0.257          & 14.59          & 15.99 \\
\textbf{2} & \textbf{0.204} & \textbf{0.210} & \textbf{18.22} & \textbf{19.55} \\
5          & 0.192          & 0.473          & 20.78          & 21.77 \\
10         & 0.271          & 0.347          & 21.82          & 23.13 \\
\bottomrule
\end{tabular}
\end{table}

Table~\ref{tab:runtime} reports throughput and accuracy under
different encoder intervals~$K$ on the most challenging VIODE
sequences (\textit{high} dynamic level).
At $K\!=\!1$, the full pipeline already operates at
${\sim}15$\,FPS, exceeding the 10\,Hz camera rate of both
evaluation datasets.
Setting $K\!=\!2$ improves throughput to ${\sim}19$\,FPS---a
\textbf{23\% speedup with no accuracy loss} (average ATE
shifts from 0.227\,m to 0.207\,m), validating the design
in Section~\ref{sec:prompt-to-mask}: temporally coherent
prompts make embedding reuse not only safe but effectively
free in terms of accuracy cost.
For $K \!\geq\! 5$, throughput gains saturate
($21$--$23$\,FPS) while nighttime accuracy degrades notably
(ATE: $0.210 \to 0.473$\,m at $K\!=\!5$), confirming that
low-contrast scenes narrow the safe reuse window.
We adopt $K\!=\!2$ as the default, achieving a favorable
accuracy--efficiency balance that demonstrates prompt
stabilization as both an accuracy enabler and a practical
efficiency mechanism.

\subsection{Real-World Deployment}
\label{sec:realworld}

%
%

Beyond public benchmarks, we evaluate STAG-VIO on a real outdoor
sequence to verify that the proposed perception-to-geometry
interface transfers to uncontrolled deployment conditions.

\textbf{Platform and ground truth.}\;
Data is collected with a four-wheeled mobile robot equipped with
an Intel RealSense D435i camera (1280$\times$720 at 30\,Hz,
BMI085 IMU at 200\,Hz) and a Livox Mid-360 LiDAR.
Camera--IMU extrinsics are calibrated with Kalibr~\cite{kalibr}.
Ground-truth poses are obtained by running
FAST-LIO2~\cite{fast-lio2} on the LiDAR data,
providing a reliable reference trajectory for evaluation.

\textbf{Scenario.}\;
The sequence traverses complex urban traffic containing a diverse
mix of dynamic objects---pedestrians, cars, buses, motorcycles,
and tricycles---presenting the combined challenges of object
diversity, varying scales, and frequent mutual occlusions.

\begin{figure}[!t]
    \centering
        \includegraphics[width=1.0\linewidth]{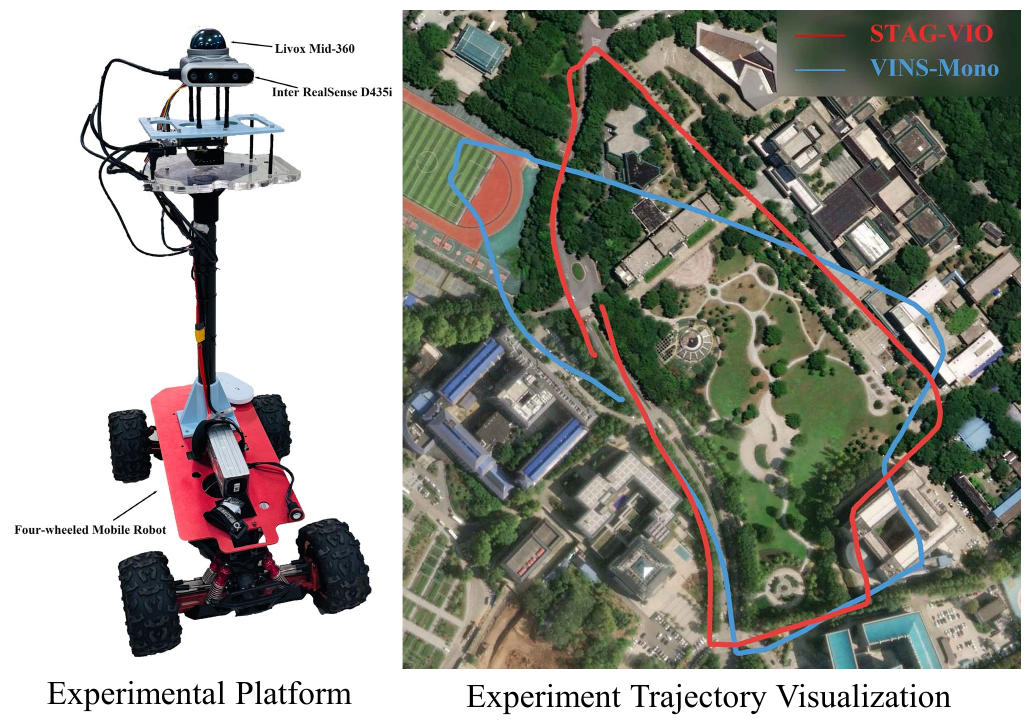}
        \caption{ (a) The sensor suite mounted on a four-wheeled mobile robot, consisting of a Livox Mid-360 LiDAR and an Intel RealSense D435i stereo camera. (b) Trajectory estimates from STAG-VIO (red) and VINS-Mono (blue) on a large-scale outdoor sequence. STAG-VIO produces a more consistent trajectory that closely follows the actual path, while VINS-Mono exhibits noticeable drift.}
        \label{fig:realexp}
        \vspace{-0.4cm}
\end{figure}

\begin{table}[!t]
\centering
\caption{Real-world experiment results.}
\label{tab:realworld}
\setlength{\tabcolsep}{5pt}
\begin{tabular}{l cccc}
\toprule
\multicolumn{1}{c}{\multirow{2}{*}{\large{Method}}}
  & \multicolumn{4}{c}{ATE [m] $\downarrow$} \\
\cmidrule(lr){2-5}
  & RMSE & Mean & Max & Std \\
\midrule
VINS-Mono~\cite{vins-mono}
  & 34.609 & 29.266 & 75.215 & 18.475 \\
\textbf{STAG-VIO (Ours)}
  & \textbf{14.563} & \textbf{12.656} & \textbf{31.736} & \textbf{7.205} \\
\midrule
\rowcolor{gray!10}
Improvement
  & \impv{57.9} & \impv{56.8} & \impv{57.8} & \impv{61.0} \\
\bottomrule
\end{tabular}
\end{table}

\textbf{Quantitative results.}\;
Table~\ref{tab:realworld} reports trajectory accuracy.
STAG-VIO reduces RMSE~ATE from 34.61\,m to 14.56\,m compared
with the VINS-Mono baseline, a \textbf{57.9\%} error reduction.
The improvement is consistent across all statistics: the maximum
error drops by 57.8\% (from 75.21\,m to 31.74\,m) and the
standard deviation decreases by 61.0\% (from 18.47\,m to
7.20\,m), indicating that STAG-VIO suppresses both average drift
and large occasional excursions.

\textbf{Qualitative results.}\;
Figure~\ref{fig:realexp} aligns estimated trajectories with
satellite imagery.
Compared with VINS-Mono, which exhibits noticeable drift in
segments with dense moving traffic, STAG-VIO produces a visibly
more consistent path that better conforms to the road geometry,
confirming that the proposed framework transfers effectively to
real-world deployments with complex dynamics and partial
occlusions.

\section{CONCLUSION}
\label{sec:conclusion}

We have shown that dynamic robustness in visual--inertial
odometry is fundamentally an interface stabilization problem.
STAG-VIO instantiates this principle through a
\mbox{Track-to-Prompt} $\to$ \mbox{Prompt-to-Mask} $\to$
\mbox{Mask-to-Geometry} pipeline, achieving consistent
improvements on both VIODE and OpenLORIS-Scene benchmarks,
with ablation confirming prompt stabilization as the single
most impactful component (up to \textbf{83\%} error reduction)
and real-world deployment validating practical effectiveness
(\textbf{57.9\%} RMSE reduction in complex urban traffic).
Current limitations include the category-bounded upstream
detector, resolution-dependent morphological parameters, and
${\sim}19$\,FPS throughput that leaves limited headroom for
higher-frequency sensors; integrating open-vocabulary
detectors~\cite{liu2024grounding}, adaptive morphological
kernels, and TensorRT deployment are promising directions.

\section*{Acknowledgments}

This work was supported in part by the National Natural Science Foundation of
China under Grant No.~42474060, the National Key R\&D Program of China under
Grant No.~2023YFB3906101, and the Natural Science Foundation of Hubei Province
under Grant No.~2024AFD403.

We would like to express our sincere gratitude to the State Key Laboratory of
Information Engineering in Surveying, Mapping and Remote Sensing (LIESMARS),
Wuhan University, for their support with the experimental platform and
facilities.

\normalsize
\bibliography{main}

@String(CVPR= {IEEE Conf. Comput. Vis. Pattern Recog.})

@String(ICCV= {Int. Conf. Comput. Vis.})

@String(CVPR  = {CVPR})

@String(ICCV  = {ICCV})

@misc{SegNet,
   author = {Badrinarayanan, Vijay and Kendall, Alex and Cipolla, Roberto},
   title = {SegNet: A Deep Convolutional Encoder-Decoder Architecture for Image Segmentation},
   pages = {arXiv:1511.00561},
   month = {November 01, 2015},
   DOI = {10.48550/arXiv.1511.00561},
   url = {https://ui.adsabs.harvard.edu/abs/2015arXiv151100561B},
   year = {2015},
   type = {Electronic Article}
}

@article{orb-slam3,
   author = {Campos, C. and Elvira, R. and Rodríguez, J. J. G. and Montiel, J. M. M. and Tardós, J. D.},
   title = {ORB-SLAM3: An Accurate Open-Source Library for Visual, Visual–Inertial, and Multimap SLAM},
   journal = {IEEE Transactions on Robotics},
   volume = {37},
   number = {6},
   pages = {1874-1890},
   ISSN = {1941-0468},
   DOI = {10.1109/TRO.2021.3075644},
   year = {2021},
   type = {Journal Article}
}

@misc{Mask_rcnn,
   author = {He, Kaiming and Gkioxari, Georgia and Dollár, Piotr and Girshick, Ross},
   title = {Mask R-CNN},
   pages = {arXiv:1703.06870},
   month = {March 01, 2017},
   note = {open source; appendix on more results},
   DOI = {10.48550/arXiv.1703.06870},
   url = {https://ui.adsabs.harvard.edu/abs/2017arXiv170306870H},
   year = {2017},
   type = {Electronic Article}
}

@inproceedings{sam,
   author = {Kirillov, A. and Mintun, E. and Ravi, N. and Mao, H. Z. and Rolland, C. and Gustafson, L. and Xiao, T. T. and Whitehead, S. and Berg, A. C. and Lo, W. Y. and Dolla'r, P. and Girshick, R. and Ieee},
   title = {Segment Anything},
   booktitle = {IEEE/CVF International Conference on Computer Vision (ICCV)},
   series = {IEEE International Conference on Computer Vision},
   pages = {3992-4003},
   note = {Kirillov, Alexander Mintun, Eric Ravi, Nikhila Mao, Hanzi Rolland, Chloe Gustafson, Laura Xiao, Tete Whitehead, Spencer Berg, Alexander C. Lo, Wan-Yen Dolla'r, Piotr Girshick, Ross
1550-5499},
   ISBN = {979-8-3503-0718-4},
   DOI = {10.1109/iccv51070.2023.00371},
   url = {<Go to ISI>://WOS:001159644304024},
   year = {2023},
   type = {Conference Proceedings}
}

@article{Dynamic-VINS,
   author = {Liu, Jianheng and Li, Xuanfu and Liu, Yueqian and Chen, Haoyao},
   title = {RGB-D Inertial Odometry for a Resource-Restricted Robot in Dynamic Environments},
   journal = {Ieee Robotics and Automation Letters},
   volume = {7},
   number = {4},
   pages = {9573-9580},
   note = {Times Cited: 24
Liu, Jianheng/0000-0002-3124-5559; Liu, Yueqian/0000-0002-1994-6408
0
24},
   ISSN = {2377-3766},
   DOI = {10.1109/lra.2022.3191193},
   url = {<Go to ISI>://WOS:000831182500036},
   year = {2022},
   type = {Journal Article}
}

@article{VIODE,
   author = {Minoda, K. and Schilling, F. and Wüest, V. and Floreano, D. and Yairi, T.},
   title = {VIODE: A Simulated Dataset to Address the Challenges of Visual-Inertial Odometry in Dynamic Environments},
   journal = {IEEE Robotics and Automation Letters},
   volume = {6},
   number = {2},
   pages = {1343-1350},
   ISSN = {2377-3766},
   DOI = {10.1109/LRA.2021.3058073},
   year = {2021},
   type = {Journal Article}
}

@article{vins-mono,
   author = {Qin, T. and Li, P. and Shen, S.},
   title = {VINS-Mono: A Robust and Versatile Monocular Visual-Inertial State Estimator},
   journal = {IEEE Transactions on Robotics},
   volume = {34},
   number = {4},
   pages = {1004-1020},
   ISSN = {1941-0468},
   DOI = {10.1109/TRO.2018.2853729},
   year = {2018},
   type = {Journal Article}
}

@misc{vins-fusion,
   author = {Qin, Tong and Pan, Jie and Cao, Shaozu and Shen, Shaojie},
   title = {A General Optimization-based Framework for Local Odometry Estimation with Multiple Sensors},
   pages = {arXiv:1901.03638},
   month = {January 01, 2019},
   DOI = {10.48550/arXiv.1901.03638},
   url = {https://ui.adsabs.harvard.edu/abs/2019arXiv190103638Q},
   year = {2019},
   type = {Electronic Article}
}

@inproceedings{rp-vio,
   author = {Ram, K. and Kharyal, C. and Harithas, S. S. and Krishna, K. M. and Ieee},
   title = {RP-VIO: Robust Plane-based Visual-Inertial Odometry for Dynamic Environments},
   booktitle = {IEEE/RSJ International Conference on Intelligent Robots and Systems (IROS)},
   series = {IEEE International Conference on Intelligent Robots and Systems},
   pages = {9198-9205},
   note = {Ram, Karnik Kharyal, Chaitanya Harithas, Sudarshan S. Krishna, K. Madhava
Krishna, Madhava/JYO-4800-2024
2153-0858},
   ISBN = {978-1-6654-1714-3},
   DOI = {10.1109/iros51168.2021.9636522},
   url = {<Go to ISI>://WOS:000755125507029},
   year = {2021},
   type = {Conference Proceedings}
}

@INPROCEEDINGS{openloris,
  author={Shi, Xuesong and Li, Dongjiang and Zhao, Pengpeng and Tian, Qinbin and Tian, Yuxin and Long, Qiwei and Zhu, Chunhao and Song, Jingwei and Qiao, Fei and Song, Le and Guo, Yangquan and Wang, Zhigang and Zhang, Yimin and Qin, Baoxing and Yang, Wei and Wang, Fangshi and Chan, Rosa H. M. and She, Qi},
  booktitle={2020 IEEE International Conference on Robotics and Automation (ICRA)}, 
  title={Are We Ready for Service Robots? The OpenLORIS-Scene Datasets for Lifelong SLAM}, 
  year={2020},
  volume={},
  number={},
  pages={3139-3145},
  doi={10.1109/ICRA40945.2020.9196638}}

@ARTICLE{dyna-vins,
  author={Song, Seungwon and Lim, Hyungtae and Lee, Alex Junho and Myung, Hyun},
  journal={IEEE Robotics and Automation Letters}, 
  title={DynaVINS: A Visual-Inertial SLAM for Dynamic Environments}, 
  year={2022},
  volume={7},
  number={4},
  pages={11523-11530},
  doi={10.1109/LRA.2022.3203231}}

@misc{mobilesam,
   author = {Zhang, Chaoning and Han, Dongshen and Qiao, Yu and Kim, Jung Uk and Bae, Sung-Ho and Lee, Seungkyu and Hong, Choong Seon},
   title = {Faster Segment Anything: Towards Lightweight SAM for Mobile Applications},
   pages = {arXiv:2306.14289},
   month = {June 01, 2023},
   note = {First work to make SAM lightweight for mobile applications},
   DOI = {10.48550/arXiv.2306.14289},
   url = {https://ui.adsabs.harvard.edu/abs/2023arXiv230614289Z},
   year = {2023},
   type = {Electronic Article}
}

@ARTICLE{fast-lio2,
  author={Xu, Wei and Cai, Yixi and He, Dongjiao and Lin, Jiarong and Zhang, Fu},
  journal={IEEE Transactions on Robotics}, 
  title={FAST-LIO2: Fast Direct LiDAR-Inertial Odometry}, 
  year={2022},
  volume={38},
  number={4},
  pages={2053-2073},
  doi={10.1109/TRO.2022.3141876}}

@INPROCEEDINGS{orb,
  author={Rublee, Ethan and Rabaud, Vincent and Konolige, Kurt and Bradski, Gary},
  booktitle={2011 International Conference on Computer Vision}, 
  title={ORB: An efficient alternative to SIFT or SURF}, 
  year={2011},
  volume={},
  number={},
  pages={2564-2571},
  doi={10.1109/ICCV.2011.6126544}}

@INPROCEEDINGS{ANMS,
  author={Brown, M. and Szeliski, R. and Winder, S.},
  booktitle={2005 IEEE Computer Society Conference on Computer Vision and Pattern Recognition (CVPR'05)}, 
  title={Multi-image matching using multi-scale oriented patches}, 
  year={2005},
  volume={1},
  number={},
  pages={510-517 vol. 1},
  doi={10.1109/CVPR.2005.235}}

@INPROCEEDINGS{PVO,
  author={Ye, Weicai and Lan, Xinyue and Chen, Shuo and Ming, Yuhang and Yu, Xingyuan and Bao, Hujun and Cui, Zhaopeng and Zhang, Guofeng},
  booktitle={2023 IEEE/CVF Conference on Computer Vision and Pattern Recognition (CVPR)}, 
  title={PVO: Panoptic Visual Odometry}, 
  year={2023},
  volume={},
  number={},
  pages={9579-9589},
  doi={10.1109/CVPR52729.2023.00924}}

@INPROCEEDINGS{Kalibr,
  author={Furgale, Paul and Rehder, Joern and Siegwart, Roland},
  booktitle={2013 IEEE/RSJ International Conference on Intelligent Robots and Systems}, 
  title={Unified temporal and spatial calibration for multi-sensor systems}, 
  year={2013},
  volume={},
  number={},
  pages={1280-1286},
  doi={10.1109/IROS.2013.6696514}}

@ARTICLE{yolov11,
       author = {{Khanam}, Rahima and {Hussain}, Muhammad},
        title = "{YOLOv11: An Overview of the Key Architectural Enhancements}",
      journal = {arXiv e-prints},
         year = 2024,
        month = oct,
          eid = {arXiv:2410.17725},
        pages = {arXiv:2410.17725},
          doi = {10.48550/arXiv.2410.17725},
archivePrefix = {arXiv},
       eprint = {2410.17725},
 primaryClass = {cs.CV},
       adsurl = {https://ui.adsabs.harvard.edu/abs/2024arXiv241017725K}
}

@misc{efficientsam2023,
  title         = {EfficientSAM: Leveraged Masked Image Pretraining for Efficient Segment Anything},
  author        = {Xiong, Yu and Zhang, Bohao and Huang, Rui and Liang, Xiaodan and Lin, Dahua and others},
  year          = {2023},
  eprint        = {2312.00863},
  archivePrefix = {arXiv},
  primaryClass  = {cs.CV},
  url           = {https://arxiv.org/abs/2312.00863}
}

@misc{mobilesamv2,
  title         = {MobileSAMv2: Faster Segment Anything to Everything},
  author        = {Zhang, Chaoning and Han, Kang and Xu, Chang and others},
  year          = {2023},
  eprint        = {2312.09579},
  archivePrefix = {arXiv},
  primaryClass  = {cs.CV},
  url           = {https://arxiv.org/abs/2312.09579}
}

@misc{fastsam2023,
  title         = {Fast Segment Anything},
  author        = {Zhao, Xinyu and Wang, Wenguan and others},
  year          = {2023},
  eprint        = {2306.12156},
  archivePrefix = {arXiv},
  primaryClass  = {cs.CV},
  url           = {https://arxiv.org/abs/2306.12156}
}

@misc{dsslam2018,
  title         = {DS-SLAM: A Semantic Visual SLAM towards Dynamic Environments},
  author        = {Yu, Chao and Liu, Zuxin and Liu, Xinjun and Xie, Fugui and Yang, Yi and Wei, Qi and Fei, Qiao},
  year          = {2018},
  eprint        = {1809.08379},
  archivePrefix = {arXiv},
  primaryClass  = {cs.RO},
  url           = {https://arxiv.org/abs/1809.08379}
}

@misc{dynaslam2018,
  title         = {DynaSLAM: Tracking, Mapping and Inpainting in Dynamic Scenes},
  author        = {Bescos, Berta and Facil, Jose M. and Civera, Javier and Neira, Jos{\'e}},
  year          = {2018},
  eprint        = {1806.05620},
  archivePrefix = {arXiv},
  primaryClass  = {cs.CV},
  url           = {https://arxiv.org/abs/1806.05620}
}

@misc{rdvio2023,
  title         = {RD-VIO: Robust Visual-Inertial Odometry for Mobile Augmented Reality in Dynamic Environments},
  author        = {Li, Jinyu and Pan, Xiaokun and Huang, Gan and Zhang, Ziyang and Wang, Nan and Bao, Hujun and Zhang, Guofeng},
  year          = {2023},
  eprint        = {2310.15072},
  archivePrefix = {arXiv},
  primaryClass  = {cs.RO},
  url           = {https://arxiv.org/abs/2310.15072}
}

@ARTICLE{dff-vio,
  author={Luo, Nan and Hu, Zhexuan and Ding, Yuan and Li, Jiaxu and Zhao, Hui and Liu, Gang and Wang, Quan},
  journal={IEEE Transactions on Circuits and Systems for Video Technology}, 
  title={DFF-VIO: A General Dynamic Feature Fused Monocular Visual-Inertial Odometry}, 
  year={2025},
  volume={35},
  number={2},
  pages={1758-1773},
  doi={10.1109/TCSVT.2024.3482573}}

@INPROCEEDINGS{sam-slam,
  author={Chen, Xianhao and Wang, Tengyue and Mai, Haonan and Yang, Liangjing},
  booktitle={2024 8th International Conference on Robotics, Control and Automation (ICRCA)}, 
  title={SamSLAM: A Visual SLAM Based on Segment Anything Model for Dynamic Environment}, 
  year={2024},
  volume={},
  number={},
  pages={91-97},
  doi={10.1109/ICRCA60878.2024.10649185}}

@article{rajabi2023computing,
  title={Computing a Many-to-Many Matching with Demands and Capacities between Two Sets Using the Hungarian Algorithm},
  author={Rajabi-Alni, Fatemeh and Bagheri, Alireza},
  journal={Journal of mathematics},
  volume={2023},
  number={1},
  pages={7761902},
  year={2023},
  publisher={Wiley Online Library}
}

@inproceedings{RT-DETR,
  title={Detrs beat yolos on real-time object detection},
  author={Zhao, Yian and Lv, Wenyu and Xu, Shangliang and Wei, Jinman and Wang, Guanzhong and Dang, Qingqing and Liu, Yi and Chen, Jie},
  booktitle={Proceedings of the IEEE/CVF conference on computer vision and pattern recognition},
  pages={16965--16974},
  year={2024}
}

@article{RT-DETRv2,
  title={Rt-detrv2: Improved baseline with bag-of-freebies for real-time detection transformer},
  author={Lv, Wenyu and Zhao, Yian and Chang, Qinyao and Huang, Kui and Wang, Guanzhong and Liu, Yi},
  journal={arXiv preprint arXiv:2407.17140},
  year={2024}
}

@article{sam2,
  title={Sam 2: Segment anything in images and videos},
  author={Ravi, Nikhila and Gabeur, Valentin and Hu, Yuan-Ting and Hu, Ronghang and Ryali, Chaitanya and Ma, Tengyu and Khedr, Haitham and R{\"a}dle, Roman and Rolland, Chloe and Gustafson, Laura and others},
  journal={arXiv preprint arXiv:2408.00714},
  year={2024}
}

@inproceedings{liu2024grounding,
  title={Grounding dino: Marrying dino with grounded pre-training for open-set object detection},
  author={Liu, Shilong and Zeng, Zhaoyang and Ren, Tianhe and Li, Feng and Zhang, Hao and Yang, Jie and Jiang, Qing and Li, Chunyuan and Yang, Jianwei and Su, Hang and others},
  booktitle={European conference on computer vision},
  pages={38--55},
  year={2024},
  organization={Springer}
}

@article{yolov5v8v10,
  title={Yolov5, yolov8 and yolov10: The go-to detectors for real-time vision},
  author={Hussain, Muhammad},
  journal={arXiv preprint arXiv:2407.02988},
  year={2024}
}

@ARTICLE{vins-mah,
  author={Hu, Yuquan and Gardi, Alessandro},
  journal={IEEE Robotics and Automation Letters}, 
  title={VINS-Mah: A Robust Monocular Visual-Inertial State Estimator for Dynamic Environments}, 
  year={2026},
  volume={11},
  number={3},
  pages={3614-3621},
  doi={10.1109/LRA.2026.3656724}}

\newpage


\appendix

\section{Detailed Experimental Statistics}
\label{app:stats}

This appendix reports the full distributional statistics underlying two
analyses in the main text: the static-constraint-budget robustness sweep
(Section~\ref{sec:robust}, Figure~\ref{fig:heatmap}) and the
prompt-stabilization ablation (Section~\ref{sec:ablation},
Figure~\ref{fig:violin}). All statistics are computed directly from the raw
per-configuration and per-frame ATE values.

\subsection{Static-Constraint-Budget Robustness Sweep}
\label{app:robust}

We sweep the maximum feature count $N_{\max}\in[40,200]$ (step 2) and the
minimum feature distance $D_{\min}\in[5,40]$\,px (step 1), giving 2{,}916
configurations per method, and evaluate ATE RMSE on the VIODE high-dynamic
sequence. Table~\ref{tab:app_robust} reports the distribution of ATE RMSE over
the full grid; ``Configs.\ $<0.5$\,m'' and ``$>1.0$\,m'' are the fractions of
configurations falling in each range. The correlation rows give the Pearson
correlation $\rho$ between ATE RMSE and each swept parameter, with the
coefficient of determination $R^2$ (the fraction of error variance explained by
that parameter). Table~\ref{tab:app_nmax} disaggregates the mean ATE RMSE into
$N_{\max}$ bands, averaged over all $D_{\min}$ values.

\begin{table}[!t]
\centering
\caption{ATE RMSE distribution across the full $(N_{\max}, D_{\min})$ parameter
sweep (2{,}916 configurations, VIODE high-dynamic sequence). All ATE values in
meters. Best value in each row in bold.}
\label{tab:app_robust}
\begin{tabular}{lcc}
\toprule
Statistic & VINS-Mono & STAG-VIO (Ours) \\
\midrule
Median                       & 2.334          & \textbf{0.229} \\
Mean                         & 3.069          & \textbf{0.274} \\
Std.\ deviation              & 2.157          & \textbf{0.303} \\
IQR ($Q_3-Q_1$)              & 4.127          & \textbf{0.165} \\
Minimum                      & 0.113          & \textbf{0.069} \\
Maximum                      & 7.668          & \textbf{4.943} \\
\midrule
Configs.\ $<0.5$\,m          & 13.6\%         & \textbf{95.2\%} \\
Configs.\ $>1.0$\,m          & 78.9\%         & \textbf{0.8\%}  \\
\midrule
Corr.\ with $N_{\max}$ ($\rho$/$R^2$) & 0.04\,/\,0.1\%  & 0.11\,/\,1.2\% \\
Corr.\ with $D_{\min}$ ($\rho$/$R^2$) & $-0.77$\,/\,58.7\% & \textbf{$-0.06$\,/\,0.3\%} \\
\bottomrule
\end{tabular}
\end{table}

\begin{table}[!t]
\centering
\caption{Mean ATE RMSE within $N_{\max}$ bands (averaged over all $D_{\min}$
values), VIODE high-dynamic sequence. All values in meters. Best value in each
column in bold.}
\label{tab:app_nmax}
\small
\setlength{\tabcolsep}{3pt}
\begin{tabular*}{\columnwidth}{@{\extracolsep{\fill}}lcccc@{}}
\toprule
$N_{\max}$ band & $[40,80)$ & $[80,120)$ & $[120,160)$ & $[160,200)$ \\
\midrule
VINS-Mono         & 2.874 & 3.122 & 3.149 & 3.128 \\
STAG-VIO (Ours)   & \textbf{0.235} & \textbf{0.251} & \textbf{0.268} & \textbf{0.339} \\
\bottomrule
\end{tabular*}
\end{table}

\subsection{Prompt-Stabilization Ablation}
\label{app:ablation}

We compare STAG-VIO with the prompt-stabilization tracker enabled (``w/'')
against a variant that feeds raw per-frame detections directly to the segmenter
(``w/o''), on the VIODE City Day and City Night sequences.
Table~\ref{tab:app_ablation} reports the full statistics of the per-frame ATE
distributions visualized in Figure~\ref{fig:violin}: central-tendency measures
(RMSE, mean, median), spread and tail statistics (standard deviation, IQR, 95th
percentile, maximum, and the fraction of frames exceeding 1\,m), and the number
of tracked frames $n$ over which each distribution is computed. The 83.28\%
(City Day) and 63.99\% (City Night) reductions quoted in the main text
correspond to the RMSE row.\footnote{These per-frame ATE statistics are
computed on the raw estimated trajectories and therefore differ slightly in
convention from the aligned ATE RMSE reported in the main comparison table
(Table~1), which follows the standard Umeyama-aligned evaluation.}

\begin{table}[!t]
\centering
\caption{Per-frame ATE distribution statistics for the prompt-stabilization
ablation on VIODE (data underlying Figure~\ref{fig:violin}). ``w/o'' feeds raw
detections to the segmenter; ``w/'' enables the proposed stabilization tracker.
All ATE values in meters. Best value of each pair in bold.}
\label{tab:app_ablation}
\begin{tabular}{lcccc}
\toprule
 & \multicolumn{2}{c}{City Day} & \multicolumn{2}{c}{City Night} \\
\cmidrule(lr){2-3}\cmidrule(lr){4-5}
Statistic          & w/o    & w/              & w/o    & w/ \\
\midrule
RMSE               & 0.868  & \textbf{0.145}  & 0.531  & \textbf{0.191} \\
Mean               & 0.824  & \textbf{0.159}  & 0.569  & \textbf{0.194} \\
Median             & 0.622  & \textbf{0.145}  & 0.528  & \textbf{0.180} \\
Std.\ deviation    & 0.609  & \textbf{0.064}  & 0.199  & \textbf{0.097} \\
IQR ($Q_3-Q_1$)    & 0.380  & \textbf{0.054}  & 0.235  & \textbf{0.151} \\
95th percentile    & 2.373  & \textbf{0.326}  & 0.954  & \textbf{0.328} \\
Maximum            & 2.608  & \textbf{0.406}  & 1.062  & \textbf{0.583} \\
Frames $>1$\,m     & 16.6\% & \textbf{0.0\%}  & 1.1\%  & \textbf{0.0\%} \\
\midrule
Frames $n$         & 614    & 616             & 568    & 561 \\
\bottomrule
\end{tabular}
\end{table}

\end{document}